\pdfoutput=1

\documentclass[11pt]{article}

\usepackage[]{acl}

\usepackage{times}
\usepackage{latexsym}
\usepackage{xcolor}
\usepackage{multirow,hhline,graphicx,array}

\usepackage[T1]{fontenc}

\usepackage[utf8]{inputenc}

\usepackage{microtype}
\usepackage{booktabs}
\usepackage{graphicx}
\usepackage{caption}
\usepackage{subcaption}
\usepackage{amsmath}
\usepackage{hyperref}

\usepackage{pgfplots}

\usepackage{tabularx,ragged2e}

\definecolor{bigbird_color}{RGB}{46,139,87}
\definecolor{longformer_color}{RGB}{107,143,179}
\definecolor{hipool_color}{RGB}{161,116,161}
\definecolor{hipool_bar}{RGB}{147,112,219}
\definecolor{longformer_bar}{RGB}{143,188,143}
\definecolor{bar_color}{RGB}{70,130,180}

\newcommand{\xhdr}[1]{{\noindent\bfseries #1}.}

\title{HiPool: Modeling Long Documents Using Graph Neural Networks}

\author{Irene R. Li\textsuperscript{1}, Aosong Feng\textsuperscript{2}, Dragomir Radev\textsuperscript{2}, Rex Ying\textsuperscript{2} \\
\textsuperscript{1}University of Tokyo, 
\textsuperscript{2}Yale University \\
{ireneli@ds.itc.u-tokyo.ac.jp, \{aosong.feng, dragomir.radev, rex.ying\}@yale.edu}
}

\begin{document}
\maketitle

\begin{abstract}


Encoding long sequences in Natural Language Processing (NLP) is a challenging problem. Though recent pretraining language models achieve satisfying performances in many NLP tasks, they are still restricted by a pre-defined maximum length, making them challenging to be extended to longer sequences. So some recent works utilize hierarchies to model long sequences. However, most of them apply sequential models for upper hierarchies, suffering from long dependency issues. In this paper, we alleviate these issues through a graph-based method.  We first chunk the sequence with a fixed length to model the sentence-level information. We then leverage graphs to model intra- and cross-sentence correlations with a new attention mechanism. Additionally, due to limited standard benchmarks for long document classification (LDC), we propose a new challenging benchmark, totaling six datasets with up to 53k samples and 4034 average tokens' length. Evaluation shows our model surpasses competitive baselines by 2.6\% in F1 score, and 4.8\% on the longest sequence dataset. Our method is shown to outperform hierarchical sequential models with better performance and scalability, especially for longer sequences.


\end{abstract}

\section{Introduction}

Transformer-based models like BERT \cite{vaswani2017atten} and RoBERTa \cite{zhuang2021roberta} have achieved satisfying results in many Natural Language Processing (NLP) tasks thanks to large-scale pretraining \cite{vaswani2017attention}. However, they usually have a fixed length limit, due to the quadratic complexity of the dense self-attention mechanism, making it challenging to encode long sequences. 

One way to solve this problem is to adapt Transformers to accommodate longer inputs and optimize the attention from BERT \cite{feng2022diffuser, jaszczur2021sparse}. BigBird \cite{zaheer2020bigbird} applies sparse attention that combines random, global, and sliding window attention in a long sequence, reducing the quadratic dependency of full attention to linear. Similarly, Longformer \cite{beltagy2020longformer} applies an efficient self-attention with dilated windows that scale linearly to the window length. Both models can take up to 4096 input tokens. Though it is possible to train even larger models for longer sequences, they are restricted by a pre-defined maximum length with poor scalability. More importantly, they fail to capture high-level structures, such as relations among sentences or paragraphs, which are essential to improving NLP system performance \cite{zhang2018learning,zhu2019hierarchical}.

Another way is to apply a hierarchical structure to process adjustable input lengths with chunking representations for scalability on long sequences. Hi-Transformer \cite{wu2021hitrans} encodes both sentence-level and document-level representations using Transformers. ToBERT \cite{pappagari2019hierarchical} applies a similar approach that stacks a sentence-level Transformer over a pretrained BERT model.
While most of the existing work models upper-level hierarchy using \textit{sequential structures}, such as multiple layers of LSTMs \cite{sepp1997lstm} or Transformers, this may still bring the long dependency issue when the sequence gets longer. To alleviate this,  we investigate graph modeling as a novel hierarchy for upper levels. Besides, we also consider inter-hierarchy relationships using a new attention mechanism.

Our key insight is to replace the sequence-based model with a hierarchical attentional graph for long documents. We first apply a basic pretrained language model, BERT or RoBERTa, to encode local representation on document chunks with a fixed length. The number of chunks could be extended for longer sequences for better scalability. Different from other works, we apply a graph neural network (GNN) \cite{zhou2018gnn} to model the upper-level hierarchy to aggregate local sentence information. This is to alleviate the long dependency issue of the sequential model. Moreover, within such a graph structure, we propose a new heterogeneous attention mechanism to consider intra- and cross- sentence-level correlations.

Our contributions are two-fold: 1) We propose HiPool with multi-level hierarchies for long sequence tasks with a novel inter-hierarchy graph attention structure. Such heterogeneous graph attention is shown to outperform hierarchical sequential models with better performance and scalability, especially for longer sequences;
2) We benchmark the LDC (long document classification) task with better scaled and length-extended datasets. 
Evaluation shows that HiPool surpasses competitive baselines by 2.6\% in F1 score, and 4.8\% on the longest sequence dataset.  
Code is available at \url{https://github.com/IreneZihuiLi/HiPool}. 
 

\section{Model}

\begin{figure}[t!]
    \centering
    \includegraphics[width=7.5cm]{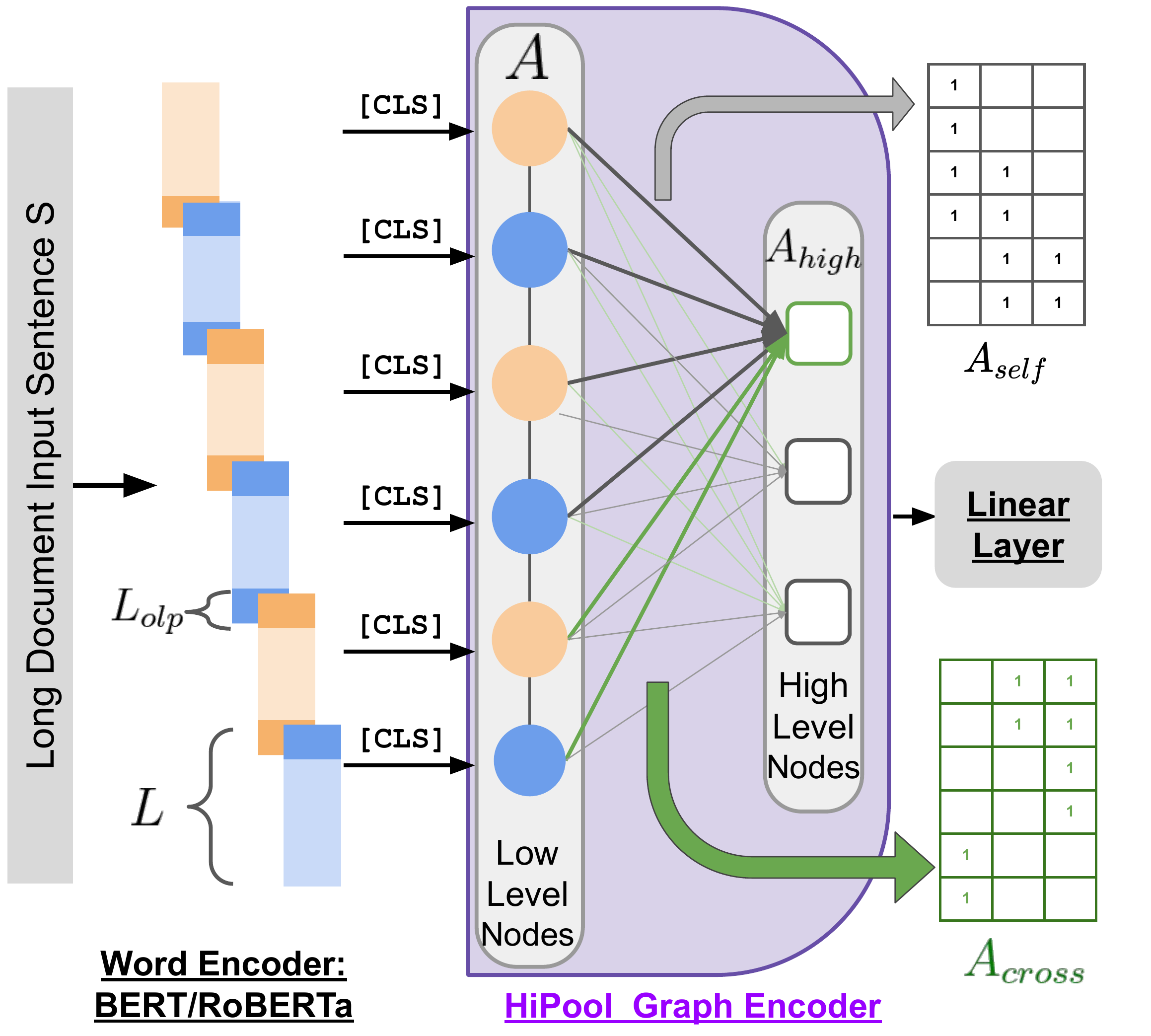}
        \caption{HiPool model illustration. It consists of a sequence encoder, HiPool graph encoder and a linear layer. }
        \label{fig:model}
\end{figure}

We introduce the HiPool (\textbf{Hi}erarchical \textbf{Pool}ing) model for long document classification, illustrated in Fig.~\ref{fig:model}. It consists of an overlapping sequence encoder, a HiPool graph encoder, and a linear layer. 

\xhdr{Overlapping Sequence Encoder} Given the input document $S$, we first chunk the document into a number of shorter pieces with a fixed length $L$, and we set the overlapping window size to be $L_{olp}$. Overlapping encoding makes it possible for a chunk to carry information from its adjacent chunks but not isolated, differentiating our model from other hierarchical ones. 
Then each chunk is encoded with a pretrained Transformer model, i.e., BERT or RoBERTa; we choose the \texttt{CLS} token representation as the input to our HiPool layer: $X = \operatorname{BERT}(S)$.



\xhdr{HiPool Graph Encoder} 
We apply a graph neural network to encode incoming word-level information. Such a model has shown its potential in some NLP tasks \cite{li-etal-2022-variational,li2021unsupervised}. 
We construct a graph, defined by $G(V,E)$, where $V$ is a set of nodes, and $E$ is a set of node connections. There are two node types: $n$ \textit{low-level nodes} and $m$ \textit{high-level nodes}, and typically $m<n$. In our experiment, we set $m=n/p$, and $p\geq 0$. The feedforward operation goes from low- to high-level nodes. In layer $l$, low-level nodes are inputs from the previous layer $l-1$, while high-level nodes at layer $l$ are computed based on low-level ones. Moreover, these high-level nodes will be the input to the next layer $l+1$, becoming the low-level nodes in that layer. We consider $X$ the low-level nodes in the first HiPool layer, as shown in the figure.



In each HiPool layer, given node representation  $H^l$ and adjacency matrix $A^l$ at layer $l$, the task is to obtain $H^{l+1}$:
\vspace{-2mm}
\begin{equation}
    H^{l+1} = \text{HiPool} (H^{l},A^{l}).
    \label{eq:graph_encoder}
\vspace{-2mm}
\end{equation}

Inspired by DiffPool \cite{ying2018diffpool}, we conduct a clustering method to aggregate information. We assign node clusters with a fixed pattern based on their position. For example, adjacent low-level neighbors should map to the same high-level clustering node. So we first define a clustering adjacency matrix $A_{self}\in {\rm I\!R}^{{n} \times {m}}$ that maps $n$ nodes to $m$ nodes, indicating the relations from low- to high- level nodes, marked as black arrows in the figure. Note that our approach allows overlapping, in which some nodes may belong to two clusters. We set the clustering sliding window to be $2p$, with a stride to be $p$. In the figure, we show the case of $p=2$. 
We denote interactions between low-level nodes by the adjacency matrix $A^l$,\footnote{We eliminated the subscript of $A_{low}$ for simplicity, and this also makes Eq.~\ref{eq:graph_encoder} more generalized as other GNNs.} and we model it using a chain graph, according to the natural order of the document.\footnote{We tested with a complete graph and BigBird attention structures but found little differences.}

Then, the relations between high-level nodes $A^l_{high}$ and their node representations $H_{high}^l$ are computed:
\vspace{-5mm}
\begin{equation}
\begin{split}
\begin{aligned}
    A^{l}_{high} &= A_{self}^T A^{l} A_{self}, \\
    H_{high}^{l} &= A_{self} H^{l}.
    \label{eq:adj}
\end{aligned}
\end{split}
\end{equation}
\vspace{-3mm}

Besides, for each high-level node, to strengthen the connections across different clusters, we propose an attention mechanism to obtain cross-sentence information. We propose a new edge type that connects external cluster low-level nodes to each high-level node, and the adjacency matrix is simply $A_{cross}=1-A_{self}$, marked by green in the figure. We update $H^{l}_{high}$ as the following:
\vspace{-2mm}
\begin{equation}
\begin{split}
\begin{aligned}
    W_{score} &= H^{l}_{self} W_{atten} (H^{l})^{T}, \\
    W_{score} &= W_{score} A_{cross}^T, \\ 
    H^{l}_{high} &\leftarrow W_{score} H^{l} + H_{high}^{l}, 
\end{aligned}
\end{split}
\end{equation}
\vspace{-3mm}

where $W_{atten}$ is trainable, and $W_{score}$ is a scoring matrix. We then apply a GNN to obtain $H^{l+1}$. For example, a graph convolution network (GCN)  \cite{kipf2016semi}: 
\vspace{-2mm}
\begin{equation}
\begin{split}
\begin{aligned}
    H^{l+1} = \text{GCN} (H^{l}_{high}, A^{l}_{high}).
\end{aligned}
\end{split}
\vspace{-5mm}
\end{equation}
We run our experiments with two layers, and apply a sum aggregator to achieve document embeddings. More HiPool layers are also possible.

\xhdr{Linear Layer}  Finally, a linear layer is connected and cross-entropy loss is applied during training. 


\section{Experiments}

\subsection{LDC Benchmark}
The LDC benchmark contains six datasets. 
We first choose four widely-used public datasets. \textbf{Hyperpartisan} (HYP) \cite{kiesel-etal-2019-semeval} and \textbf{20NewsGroups} (20NG) \cite{lang1995news} are both news text datasets with different scales.  \textbf{IMDB} \cite{maas-etal-2011-learning} is a movie review dataset for sentiment classification. \textbf{ILDC} \cite{malik-etal-2021-ildc} is a large corpus of legal cases annotated with binary court decisions (\lq\lq accepted\rq \rq and \lq\lq rejected\rq \rq).  

\xhdr{Limitation and new datasets} However, 20NewsGroups and IMDB cannot test the limit of models in encoding long documents since the average length of sentence is still relatively small; whereas Hyperpartisan only contains 645 examples and is thus prone to overfitting and not representative. ILDC is large and contains long texts, but it is mainly in the legal domain. Therefore, to enrich evaluation scenario, we select and propose two new benchmarks with longer documents based on an existing large-scale corpus, Amazon product reviews \cite{he2016modeling}, to conduct long document classification. \textbf{Amazon-512} (A-512) contains all reviews that are longer than 512 words from the \textit{Electronics} category; \textbf{Amazon-2048} (A-2048) contains 10,000 randomly sampled reviews that are longer than 2048 words from the \textit{Books} category. We randomly split 8/1/1 as train/dev/test sets for both datasets. 
The proposed datasets enable us to draw statistically significant conclusions on model performance as sequence lengths increase, as demonstrated in in Table~\ref{tab1:data}.

\begin{table}[t]
\small
\resizebox{.5\textwidth}{!}{%
\begin{tabular}{lrrrrrr} \toprule
           & \textbf{HYP} & \textbf{20NG} & \textbf{IMDB}     & \textbf{A-512} & \textbf{A-2048} & \textbf{ILDC}\\ \midrule
Mean       & 741.44   & 587.56  & 301.14 & 879.62 &  2,915.03 &4039.85\\
Max        & 5,368          & 144,592       & 3,152     & 17,988     &  14,120 & 501,091\\
Min        & 21            & 37           & 10       & 512   &   2,048  & 53 \\
Med.        & 547           & 360          & 225      & 725    &  2,505  & 2,663 \\
95pt. & 2,030        & 1,229         & 771      & 1,696  &  5,216   & 11,416 \\ \midrule
Total      & 645           & 18,846        & 50,000    & 53,471 &  10,000  & 34,816 \\ 
Class      & 2             & 20           & 2        & 5  & 5    & 2   \\ \bottomrule
\end{tabular}%
}
\vspace{-1mm}
\caption{Dataset statistics on LDC benchmark. \texttt{Med.} is the median value. \texttt{95pt.} indicates 95th percentile. \texttt{Class} indicates the number of classes.}
\vspace{-6mm}
\label{tab1:data}
\end{table}

\begin{table*}[t]
\centering
\resizebox{.99\textwidth}{!}{%
\begin{tabular}{l|cccccc|c} \toprule
             & \textbf{HYP}          & \textbf{20NG}         & \textbf{IMDB}         & \textbf{A-512}        &\textbf{A-2048}   &\textbf{ILDC}    & \textbf{Avg.}   \\ \midrule
BERT    & 0.857        & 0.853        & 0.913        & 0.592        & 0.503    & 0.556   & 0.712  \\
RoBERTa & 0.874        & 0.857        & 0.953        & 0.650        & 0.579    & 0.560	& 0.745  \\
BigBird      & 0.922        & 0.823        & 0.952        & 0.674   & 0.636    & 0.637 &	\underline{0.774}  \\
Longformer    & \textbf{0.938}        & 0.863        & \textbf{0.957}        & 0.673        & 0.612     &  0.562	& 0.768  \\
ToBERT       & 0.862        & 0.901        & 0.924        & 0.587        & 0.560       & 0.611	& 0.741 \\ \midrule
HiPool-\small{BERT}     & 0.865\small{±0.030} & \textbf{0.908\small{±0.005}} & 0.931±\small{0.001} & 0.660±\small{0.009} & 0.612±\small{0.011} & 0.651±\small{0.010} &	0.771 \\
HiPool-\small{RoBERTa}  & 0.886±\small{0.018} & 0.904±\small{0.001} & 0.948±\small{0.001} & \textbf{0.690±\small{0.007}} & \textbf{0.648±\small{0.017}} & \textbf{0.685±\small{0.018}} &	\textbf{0.794}\\ \bottomrule
\end{tabular}%
}

\vspace{-2mm}
\caption{Main evaluation results on LDC benchmark. We underscore the best average of baselines, and bold the best overall models. }
\label{tab:main}
\vspace{-5mm}
\end{table*}


\subsection{Evaluation}
\xhdr{Hyperparameters} We list details in Appendix \ref{app:hyper}. 

\xhdr{Baselines} We select four pretrained models: BERT \cite{devlin-etal-2019-bert}, RoBERTa \cite{zhuang2021roberta}, BigBird \cite{zaheer2020bigbird} and Longformer \cite{beltagy2020longformer}. We also compare with a hierarchical Transformer model ToBERT \cite{pappagari2019hierarchical}. 
Hi-Transformer \cite{wu2021hitrans} failed to be reproduced as there is no code available. 
We evaluate two variations of our HiPool method by changing the sequence encoder model: HiPool-BERT and HiPool-RoBERTa. We report the Micro-F1 score in Tab.~\ref{tab:main}. 

\xhdr{Main Results} Among the pretrained models, Longformer and BigBird perform better than BERT and RoBERTa.
ToBERT can only surpass BERT as it is a hierarchical model that applies BERT as its text encoder. On average, HiPool-BERT improves significantly on BERT by 5.9\% and on ToBERT by 3\%.
Compared to ToBERT, the superior performance of HiPool can be explained by the fact that sentence-level representations in ToBERT fails to capture cross-sentence information.
HiPool surpasses baselines on A-512, A-2048 and ILDC that contain longer sequences. Notably, the best model, HiPool-RoBERTa, outperforms BigBird by 4.8\% on ILDC.
While our model applies a basic pretrained text encoder (the maximum length is 512), it can still surpass larger pretrained language models (i.e., the maximum length is 4096). 
Although HiPool is worse on HYP and IMDB, we note that HYP only has 65 examples in testing and is prone to overfitting. We further show that even in IMDB, HiPool still out-performs the best model for long sequence in Appendix \ref{app:imdb}.

\textbf{Hierarchy variations.} 
To further compare sequential and graph hierarchy, we keep the word encoder and replace the HiPool graph encoder with the following sequential modules: \texttt{Simple} linear summation over low-level nodes; \texttt{CNN} applies a 1-dimension convolution; \texttt{Trans} is to apply a Transformer on top of low-level nodes. Besides, we also look at multiple graph settings:  \texttt{Aggr-mean} is to use a mean aggregator to obtain the final document representation; \texttt{Aggr-std} is to use a feature-wise standard deviation aggregator; finally, \texttt{Aggr-pcp} applies Principal Neighbourhood Aggregation (PNA) \cite{corso2020principal}. We report results on Amazon-2048 in Tab.~\ref{tab:hier}, as it has the longest sequence on average. An observation is that applying aggregators are better than simpler structures, while keeping a graph is still a better choice. HiPool also considers attention in message passing, so it is doing even better. We also test other variations in Appendix \ref{app:graph_variations}.

\begin{table}[t]
\small
\centering
\begin{tabular}{lc|lc}
\toprule
\textbf{Hierarchy}         & \textbf{F1 } & \textbf{Hierarchy}         & \textbf{F1 }   \\ \midrule
\textit{Sequential} && \textit{Graph} & \\
Simple            & 0.618 & Aggr-mean         & 0.621 \\
CNN            & 0.608 & Aggr-std          & 0.620 \\
Trans.   & 0.560 & Aggr-pna    & 0.633 \\

&&HiPool & \textbf{0.648} \\
\bottomrule
\end{tabular}
\caption{Comparison of multiple hierarchies.}
\label{tab:hier}
\vspace{-5mm}
\end{table}

\subsection{Ablation Study}
\begin{figure}
\centering
\begin{subfigure}[b]{.98\linewidth}
\begin{tikzpicture}
\centering
\tikzstyle{every node}=[font=\small]
\begin{axis}[
    width  = 0.95\linewidth,
    height = 0.6\linewidth,
    ymin = 0.43, ymax = 0.68,
    symbolic x coords = {64, 256, 512, 1024, 2048, 4096},
    ytick = {},
    ylabel = {F1},
    ymajorgrids = true,
    xmajorgrids = true,
    y label style={at={(0.05,0.5)}},
    enlarge x limits={abs=0.3cm},
     legend style={
        at={(0.65,0.6)},
        anchor=north west,
        nodes={scale=0.5, transform shape},
        }
]
\addplot[color = bigbird_color, mark = star ]
    coordinates {
    (64, 0.484)
        (256, 0.541)
      (512, 0.592)
      (1024, 0.617)
      (2048, 0.635)
      (4096, 0.636)
    };
\addlegendentry{BigBird}

\addplot[color = blue, mark = star]
    coordinates {
    (64, 	0.486)
    (256,0.548)
      (512, 0.552)
      (1024, 0.596)
      (2048, 0.612)
      (4096, 0.594)
    };
\addlegendentry{Longformer}

\addplot[color = purple, mark = star]
    coordinates {
    (64, 0.487)
      (256, 0.546)
      (512, 0.613)
      (1024, 0.627)
      (2048, 0.6482)
      (4096, 0.6524)
    };
\addlegendentry{HiPool}
\end{axis}
\end{tikzpicture}
\caption{Amazon-2048}
\label{fig:tokenlength_amazon}

\end{subfigure}
\begin{subfigure}[b]{.98\linewidth}
\begin{tikzpicture}
\centering
\tikzstyle{every node}=[font=\small]
\begin{axis}[
    width  = 0.95\linewidth,
    height = 0.6\linewidth,
    ymin = 0.48, ymax = 0.75,
    symbolic x coords = {64, 256, 512, 1024, 2048, 4096},
    ytick = {},
    ylabel = {F1},
    ymajorgrids = true,
    xmajorgrids = true,
    y label style={at={(0.05,0.5)}},
    enlarge x limits={abs=0.3cm},
     legend style={
        at={(0.35,0.4)},
        anchor=south east,
        nodes={scale=0.5, transform shape},
        }
]
\addplot[color = bigbird_color, mark = star ]
    coordinates {
    (64, 0.5023071852)
        (256, 0.5234014502)
      (512, 0.5293342123)
      (1024, 0.5642715887)
      (2048, 0.644693474)
      (4096, 0.6374423204)
    };
\addlegendentry{BigBird}

\addplot[color = blue, mark = star]
    coordinates {
    (64, 	0.5471324984)
    (256,0.5563612393)
       (512, 0.5385629532)
      (1024, 0.5728411338)
      (2048, 0.5754779169)
      (4096, 0.5616348055)
    };
\addlegendentry{Longformer}

\addplot[color = purple, mark = star]
    coordinates {
    (64, 0.566249176)
       (256, 0.5787738958)
       (512, 0.5807514832)
      (1024, 0.5886618326)
      (2048, 0.6255767963)
      (4096, 0.6901779829)
    };
\addlegendentry{HiPool}
\end{axis}
\end{tikzpicture}
\caption{ILDC}
\label{fig:tokenlength_ildc}
\end{subfigure}
\caption{Ablation study on the input text length. (X-axis shows input length.)}
\label{fig:tokenlength}
\end{figure}
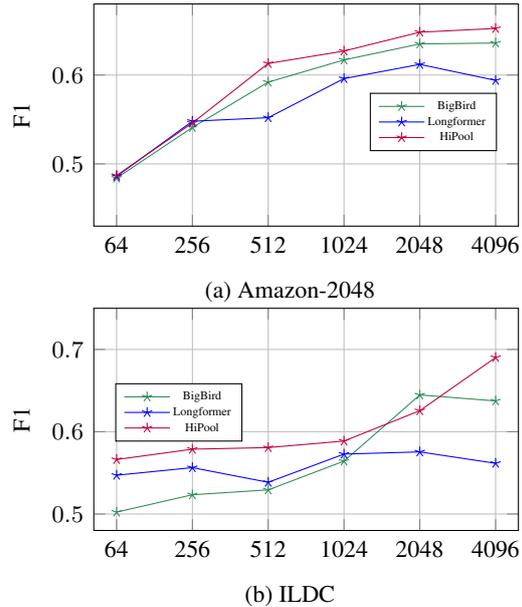

\xhdr{Effect of input length} 
To better understand the effect of input length, in Fig.~\ref{fig:tokenlength}, we present an ablation study on the Amazon-2048 and ILDC, and compare three models: BigBird, Longformer, and HiPool. In general, the models benefit from longer input sequences in both datasets. 
Interestingly, when sequence is larger than 2048, Longformer and Bigbird could not improve and they are limited in maximum lengths. In contrast, as the input sequence gets longer, HiPool steadily improves, showing its ability to encode long documents in a hierarchical structure.

\begin{table}[t]
\centering
\small
\begin{tabular}{l|rrr|r} \toprule
                 & \textbf{A-512} & \textbf{A-2048} & \textbf{ILDC} & \textbf{Avg.} \\ \midrule
HiPool-\small{RoBERTa} & \textbf{0.690} &	\textbf{0.648} &	\textbf{0.685} & \textbf{0.674} \\
\quad w/o RoBERTa      & 0.660   & 0.612                             & 0.651   & 0.641       \\
\quad w/o HiPool    & 0.601                              & 0.578                               & 0.620                             & 0.600         \\
\quad  w/o Overlapping & 0.587                              & 0.560                               & 0.611                             & 0.586        \\ \bottomrule
\end{tabular}
\caption{The effect of sequence encoding layer, HiPool layer and overlapping modules.}
\label{tab_module}
\end{table}



\xhdr{Model component} Next, we look at how each component of HiPool affects performance. As shown in Tab.~\ref{tab_module}, we first take the best model setting, HiPool-RoBERTa, and compare it with the following settings: 1) \texttt{w/o RoBERTa} is to replace RoBERTa with BERT, then the model becomes HiPool-BERT; 2) \texttt{w/o HiPool} is to remove the proposed HiPool module and replace with a simple CNN \cite{kim-2014-convolutional}; 3) \texttt{w/o Overlapping} is to remove the overlapping word encoding. We could see that removing the HiPool Layer leads to a significant drop, indicating the importance of the proposed method. Moreover, the HiPool framework can work with many pretrained language models, as we can see that applying RoBERTa improves BERT. 
A complete result table can be found in Appendix.



\section{Conclusion}
In this paper, we proposed a hierarchical framework for long document classification. The evaluation shows our model surpasses competitive baselines. 

\section{Limitations and Potential Risks}

\textbf{Limitations} 
The model we proposed is specifically for classification, while it is possible to be extended to other NLP tasks by changing the high-level task-specific layer. Besides, in the evaluation, we focused on English corpora. We plan to test on other languages in the future.

\textbf{Potential Risks}
We make our code publicly available so that everyone can access our code.  As the model is a classification model, it does not generate risky content. Users should also notice that the classification predictions may not be perfectly correct.

\section{Acknowledgements}

This paper is dedicated to the memory of Professor Dragomir Radev, who passed away while this paper was being peer-reviewed.
\begin{center}
  \includegraphics[width=0.4\linewidth]{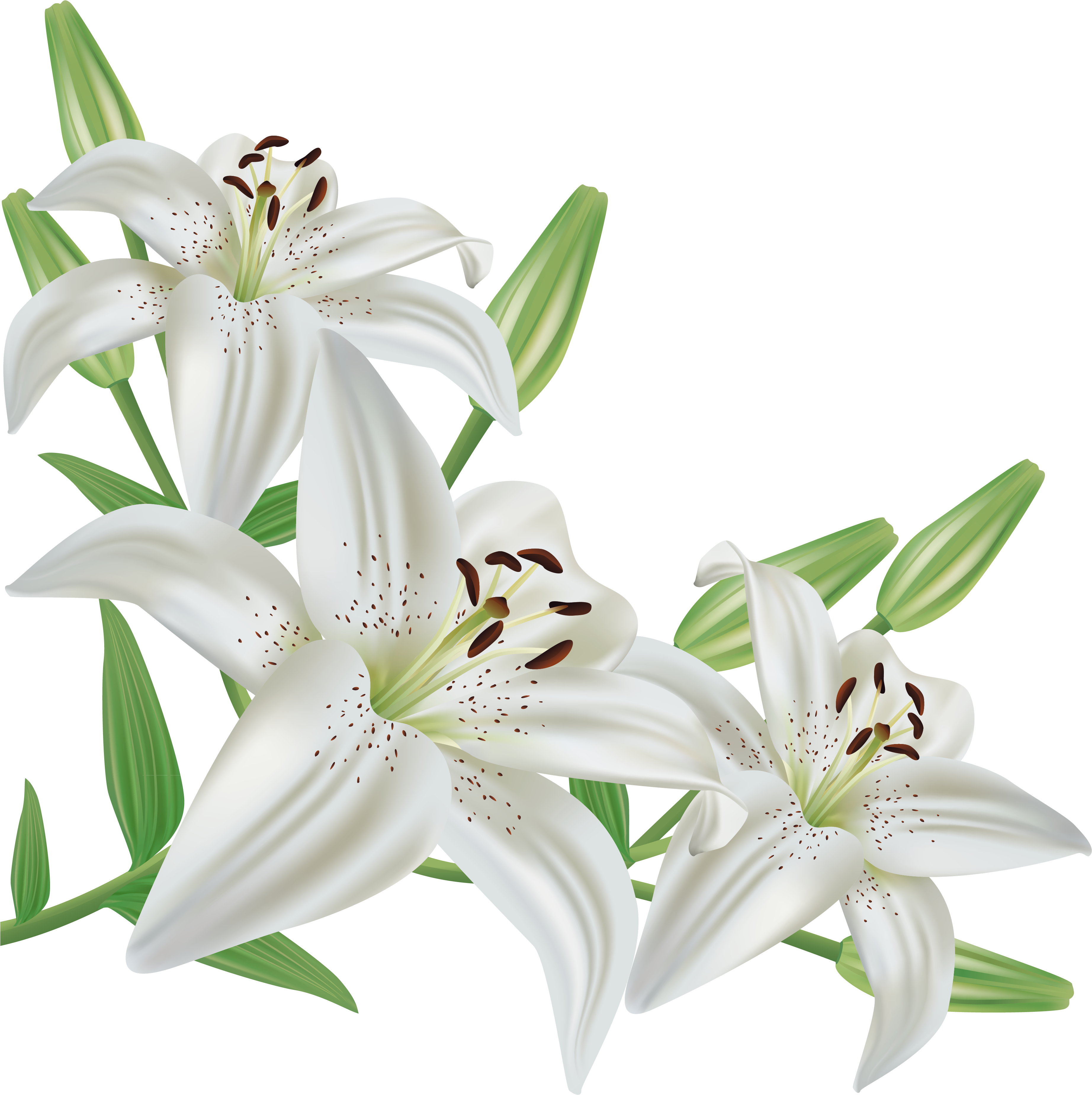}
\end{center}

\bibliography{anthology,custom}
\bibliographystyle{acl_natbib}

\appendix

\onecolumn
\section{IMDB-long Dataset}
\label{app:imdb}
\begin{figure}[h!]
\centering
\small
\begin{tikzpicture}
\begin{axis}[
    ybar,
    enlarge x limits={abs=1cm},
    bar width=0.6cm,
    legend pos = north east, 
    ymajorgrids = true,
    ylabel={F1},
    symbolic x coords={BigBird, Longformer, HiPool},
    xtick=data,
    height = 5 cm,
    width  = 0.4\textwidth,
    xtick pos=left,
    ytick pos=left,
    ]
\addplot[fill=bar_color, draw=none] coordinates {(BigBird, 0.9221538462) (Longformer, 0.9264615385) (HiPool, 0.9289230769)};

\end{axis}
\end{tikzpicture}
\caption{Performance on IMDB-long. HiPool outperforms BigBird and Longformer when the sequence length is larger than 512. }
\label{fig:IMDB-long}

\end{figure}
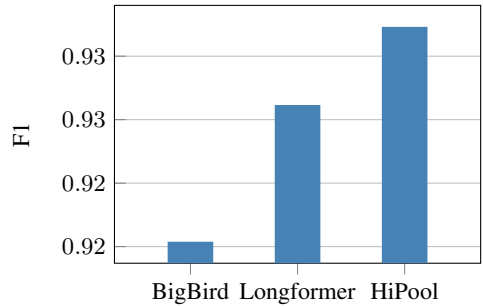

\textbf{HiPool Performs The Best for Long Sequences in IMDB.}  
As a supplementary analysis, we look at the IMDB dataset, in which HiPool performs worse than BigBird and Longformer. We filter out the sequences that are longer than 512 tokens to construct the \textbf{IMDB-long} dataset, resulting in 3250 and 3490 samples for training and testing. We show the detailed statistics of the IMDB-long dataset in Tab.~\ref{tab:imdb-long}. We show the evaluation in Fig.~\ref{fig:IMDB-long}. We can observe that HiPool can do better for long sequences.

\begin{table*}[th!]
\centering
\begin{tabular}{lrr} \toprule
     & \textbf{Train}  & \textbf{Test}   \\ \midrule
Mean       & 761.35 & 764.65 \\
Max        & 2,977  & 3,152  \\
Min        & 512    & 512    \\
Med        & 689    & 693    \\
50th pctl. & 689    & 693    \\
95th pctl. & 1,236  & 1,232  \\
Total      & 3,250  & 3,490  \\ \bottomrule
\end{tabular}
\caption{IMDB-long dataset statistics.}
\label{tab:imdb-long}
\end{table*}

\section{Graph Variations}
\label{app:graph_variations}
We study other possible GNN types for hierarchy modeling. In Eq.~\ref{eq:graph_encoder}, we replace the HiPool graph encoder with a GCN or GAT encoder. We apply two layers of the graph networks before the linear layer to compare fairly, and show results in Fig.~\ref{tab:gnn_bert_roberta}. We notice that using GCN and GAT results in lower performance than that of HiPool. A possible reason is that they only focus on modeling the low-level nodes, ignoring a cross-sentence attention mechanism to strengthen high-level communication on long sequences like HiPool. 


\begin{table*}[th!]
\centering
\begin{tabular}{l|cccccc|c} \toprule
               & \textbf{HYP}   & \textbf{20NG}  & \textbf{IMDB}  & \textbf{A-512} & \textbf{A-2048} & \textbf{ILDC} & \textbf{Avg.}  \\ \midrule

BERT-GCN       & 0.859 & 0.904 & 0.927 & 0.645 & 0.591  & 0.623 & 0.758 \\
BERT-GAT       & 0.846 & 0.907 & 0.929 & 0.653 & 0.602  & 0.626 & 0.760 \\
BERT-HiPool    & 0.865 & 0.908 & 0.931 & 0.660 & 0.612  & 0.651 & \textbf{ 0.771} \\ \midrule
RoBERTa-GCN    & 0.874 & 0.903 & 0.944 & 0.670 & 0.631  & 0.656 & 0.780 \\
RoBERTa-GAT    & 0.849 & 0.899 & 0.945 & 0.678 & 0.640  & 0.673 & 0.781 \\
RoBERTa-HiPool & 0.886 & 0.904 & 0.948 & 0.690 & 0.648  & 0.690 & \textbf{0.794} \\ \bottomrule

\end{tabular}
\caption{Comparison of other GNN types: we report F1 scores for individual dataset and the average. HiPool method is better than GCN and GAT. }
\label{tab:gnn_bert_roberta}
\end{table*}


\section{Hyperparameters, Experimental Settings}
\label{app:hyper}
We run our experiments on 4 NVIDIA RTX A6000 GPUs, with the memory to be 48GB. We list hyperparameters for baselines and HiPool model in Tab.~\ref{tab:hyperpara}. For all datasets, we apply Adam optimizer \cite{kingma2014adam} for all experiments. For HiPool, we set the chunk length $L=300$, and the overlapping length $L_{olp}$ is $L/2=150$. We apply two layers of HiPool, reducing the number of nodes for each layer by $p=2$. Among the baseline models, ToBERT \cite{pappagari2019hierarchical} is adjustable for the maximum length, because it takes the maximum value in a batch during training. We evaluated F1 scores using scikit-learn: \url{https://scikit-learn.org/stable/}.

\begin{table}[h!]
\centering
\begin{tabular}{lccccccc} \toprule
                             & \textbf{HYP}  & \textbf{20NG} & \textbf{IMDB} & \textbf{A-512} & \textbf{A-1024} & \textbf{ILDC} & \textbf{Time*}\\ \midrule
\multicolumn{7}{l}{\textit{BERT, RoBERTa}}  & 20  \\
max\_len                     & 512 & 512 & 512 & 512  & 512 & 512  & \\
\#epoch                      & 10   & 10   & 10   & 10    & 10 & 10   & \\
learning rate                & 5e-6 & 5e-6 & 5e-6 & 5e-6  & 5e-6   & 5e-6 & \\ \midrule
\multicolumn{7}{l}{\textit{BigBird, Longformer}}  & 40  \\
max\_len                     & 1024 & 1024 & 1024 & 2048  & 4096  & 4096 &  \\
\#epoch                      & 10   & 10   & 10   & 10    & 10  & 10    & \\
learning rate                & 5e-6 & 5e-6 & 5e-6 & 5e-6  & 5e-6  & 5e-6 & \\ \midrule
\multicolumn{7}{l}{\textit{ToBERT}}  &  25 \\
\#epoch                      & 8    & 10   & 10   & 12    & 12    & 12  & \\
learning rate               & 1e-5 & 1e-5 & 1e-5 & 1e-5  & 1e-5   & 1e-5  & \\\midrule
\multicolumn{7}{l}{\textit{HiPool}}    &     50$\times$5       \\
\#max\_node                  & 10   & 8    & 8    & 10    & 15    & 15   & \\
\#epoch                      & 8    & 10   & 10   & 12    & 12     & 12  & \\
learning rate: BERT                     & 1e-5 & 1e-5 & 1e-5 & 1e-5  & 1e-5 & 5e-6  &  \\
learning rate: RoBERTa                  & 5e-6 & 5e-6 & 5e-6 & 5e-6  & 5e-6   & 5e-6 & \\ \bottomrule
\end{tabular}
\caption{Hyperparameters for baseline models and HiPool. \texttt{Time*} indicates how many hours on overall trial, training and testing using a single GPU. Note that we report average and standard deviation for HiPool, so we ran the evaluation at least 5 times there.}
\label{tab:hyperpara}
\end{table}

\section{Frequently Asked Questions}
\begin{itemize}
   
    \item Q: \textit{Why do we call it a heterogeneous graph?}
    
    A: We use the term \lq \lq heterogeneous\rq\rq to distinguish the nodes from the graph. We wish to emphasize that the nodes are not the same, and they come from multiple levels and represent different information. 
    
    \item Q: \textit{Are there other possible variations for modeling the hierarchy?}
    
    A: Yes, our HiPool model is a framework that applies a graph structure for high-level hierarchy, so it is possible to apply other GNN models. One can use Relational Graph Convolutional Networks (R-GCNs) \cite{rgcn2018} to model the different relations for $A_{self}$ and $A_{cross}$. Besides, some inductive methods like GraphSAGE \cite{graphsage} can also be applied to obtain node embeddings in the graph. We leave this topic as future work. 
    
    \item Q: \textit{How does the aggregator work in Tab.~\ref{tab:hier}.?}
    
    A: We replace the sum aggregator of our original HiPool with those mentioned aggregators. The applied PyTorch implementation: \url{https://pytorch-geometric.readthedocs.io/en/latest/modules/nn.html#aggregation-operators}. 
    
     \item Q: \textit{Why did not evaluate on the LRA (Long Range Arena) \cite{lra2021tay} benchmark? }
    
    A: LRA is more suitable for testing the efficiency of Transformer-based models and it consists of multiple types of long sequences. As we mentioned in the Introduction, our proposed model belongs to another category for long sequence encoding, not the efficiency transformer category that focuses on optimizing $KQV$ attention.

\end{itemize}




\end{document}